\documentclass[journal]{IEEEtran}
\usepackage{hyperref}
\usepackage{graphicx}
\usepackage{amsmath}
\usepackage{subcaption}
\usepackage{changepage} 
\usepackage{multirow}

\ifCLASSINFOpdf
\else
\fi
\begin{document}
%
\title{
Enhancing Skin Disease Classification Leveraging Transformer-based Deep Learning Architectures and Explainable AI}

%
%
%

\author{Jayanth Mohan\IEEEauthorrefmark{1}\IEEEauthorrefmark{3}, Arrun Sivasubramanian\IEEEauthorrefmark{1}\IEEEauthorrefmark{3}, Sowmya V\IEEEauthorrefmark{1}, Vinayakumar Ravi\IEEEauthorrefmark{2}

\IEEEauthorblockA{\IEEEauthorrefmark{1}Amrita School of Artificial Intelligence, Coimbatore, Amrita Vishwa Vidyapeetham, India.\\
\IEEEauthorrefmark{2}Center for Artificial Intelligence, Prince Mohammed Bin Fahd University, Khobar, Saudi Arabia\\
\IEEEauthorrefmark{3}Equal Contribution\\
}

\thanks {Email Addresses: Jayanth Mohan (jay.thinkai@gmail.com), Arrun Sivasubramanian (arrun.sivasubramanian@gmail.com), Sowmya V (v\_sowmya@cb.amrita.edu), Vinayakumar Ravi (vravi@pmu.edu.sa)
}}



%
%

\markboth{}%
{Shell \MakeLowercase{\textit{et al.}}: Jayanth Mohan and Arrun Sivasubramanian}
%



\maketitle

\begin{abstract}
Skin diseases affect over a third of the global population, yet their impact is often underestimated. Automating skin disease classification to assist doctors with their prognosis might be difficult. Nevertheless, due to efficient feature extraction pipelines, deep learning techniques have shown much promise for various tasks, including dermatological disease identification. This study uses a skin disease dataset with 31 classes and compares it with all versions of Vision Transformers, Swin Transformers and DivoV2. The analysis is also extended to compare with benchmark convolution-based architecture presented in the literature. Transfer learning with ImageNet1k weights on the skin disease dataset contributes to a high test accuracy of 96.48\% and an F1-Score of 0.9727 using DinoV2, which is almost a 10\% improvement over this data's current benchmark results. The performance of DinoV2 was also compared for the HAM10000 and Dermnet datasets to test the model's robustness, and the trained model overcomes the benchmark results by a slight margin in test accuracy and in F1-Score on the 23 and 7 class datasets. The results are substantiated using explainable AI frameworks like GradCAM and SHAP, which provide precise image locations to map the disease, assisting dermatologists in early detection, prompt prognosis, and treatment.
\end{abstract}


%
\IEEEpeerreviewmaketitle

\section{Introduction}
%
%
%
%
\IEEEPARstart{H}{uman} skin serves various functions, including protecting the human body from contaminants, heat, and UV radiation \cite{ref1}. Skin disorders are significantly more common than we know, with impairment from skin and subcutaneous diseases accounting for 4.02\% of the total disability cases in India in 2017 \cite{ref2}. These skin diseases are growing increasingly hazardous as time passes. Dermatologists believe it can be addressed if the injury is recognized in time, but things might get tricky when they rely on manual approaches alone to identify diseases. The fundamental reason for this is that there are many types of diseases. Furthermore, physical diagnosis might be challenging because many skin diseases have similar visual characteristics that further increase difficulty in diagnosis and suggesting medical treatment \cite{ref3}.

The severity and symptoms of these skin issues vary greatly, with some skin diseases being hereditary while outside influences cause others. Over 3000 acute and chronic skin disorders affecting persons of various ages and genders have been recorded \cite{ref5}. They might be temporary or permanent and can be unpleasant or lethal in a few cases, like melanoma. Though they can be treated with medication, lotions, ointments, or lifestyle modifications \cite{ref4}, they can significantly burden patients through decreased quality of life, confidence, and higher costs. 

Deep learning (DL) techniques, especially convolutional neural networks (CNNs), have been essential in unsupervised feature extraction from images in recent years \cite{ref6}. Many academics have created many CNN designs to improve the performance in domains that have high availability and diverse annotated data \cite{ref7}, and they have also played an essential part in medical image-based classification and analysis \cite{ref8,ref9}. In the big data era, high-performance GPUs have also enabled mapping a big dataset on a network for improved CNN implementation \cite{ref10}. All these factors have helped reduce human error and variability in medical diagnoses, leading to improved patient safety and satisfaction, while enhancing diagnostic efficiency and accuracy,

Following extraordinary success on natural language tasks, transformer neural networks have been effectively applied to various computer vision challenges, yielding state-of-the-art results and pushing academics to reassess the dominance of CNNs \cite{ref11}. Taking advantage of developments in computer vision, the medical imaging profession has seen increased interest in transformers that can capture global context as opposed to CNNs with local receptive fields \cite{ref12,ref13}. Though works \cite{ref14,ref15,ref34} have explored transformers for SDC, their study is limited to models that classify skin diseases for a small corpus. They are also trained on data containing samples belonging to fewer classes, which limits the diversity of the diseases in the study. As demonstrated in our work, the models solely used by them cannot capture diversity in the distribution of diseases, and the introduction of transformer architectures such as DinoV2 in the computer vision community warrants its utilization for complex and important dermatological tasks such as SDC, which could help the general public as well as dermatologists in terms of time and resources. Moreover, the works do not provide any insight into the extent of the spread of the disease that could further help determine factors like severity or rate of spread of the disease.

\begin{figure*}[h]
\centering
\includegraphics[width=\textwidth]{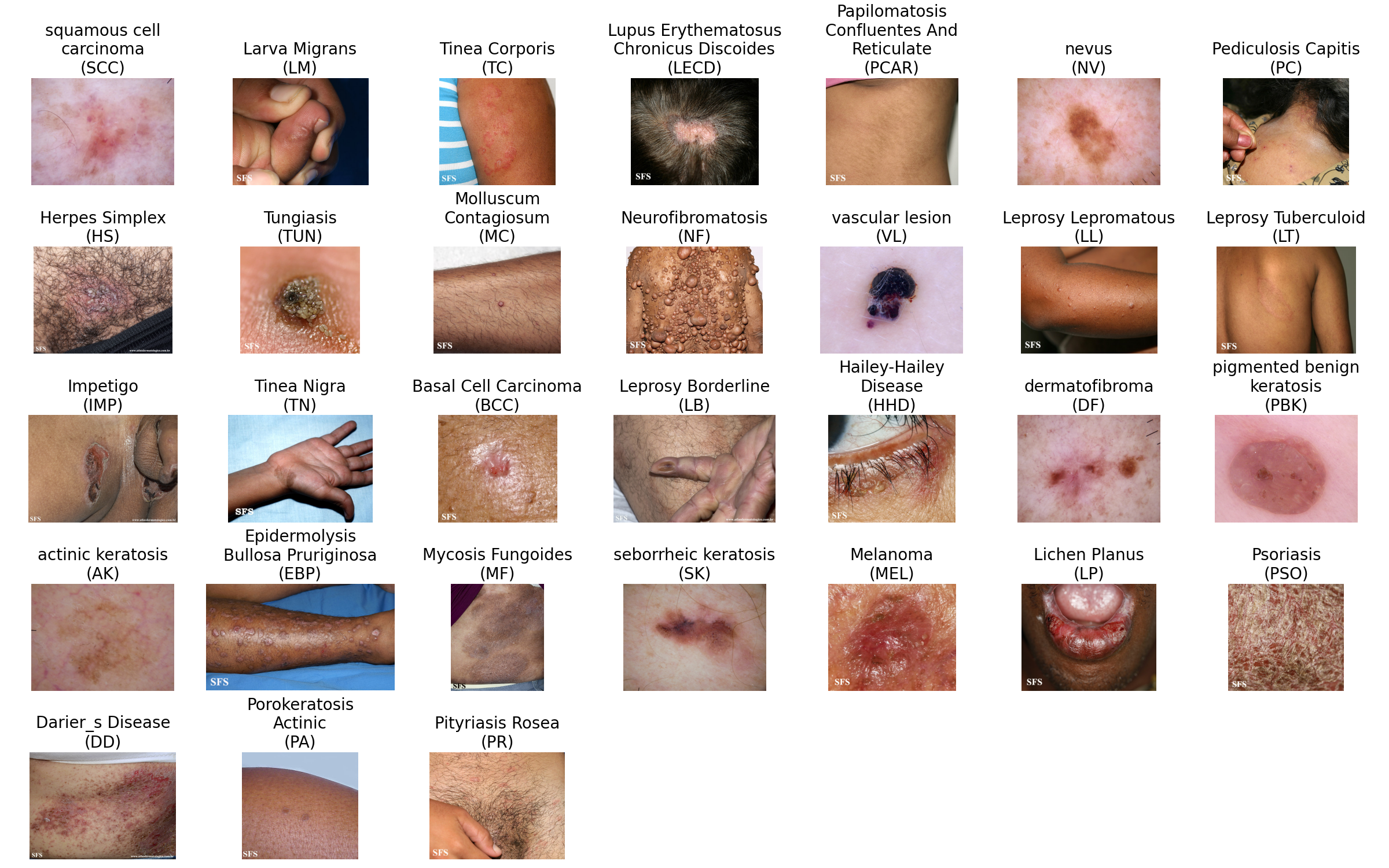}
\caption{Sample images of each of the 31 classes (with abbreviations) of the SDC dataset \cite{ref26}.}
\label{fig1}
\end{figure*}

Thus, this work addresses leveraging transformer architectures, such as Vision Transformers (ViT), Swin Transformers, and DinoV2, to classify a diverse list of skin diseases. All the variants of these models are trained and tested on a dataset containing 31 skin diseases and their augmented versions to overcome regularization and assist with data-limited classes to perform a comprehensive analysis. The samples of each class of the overall dataset are shown in Figure \ref{fig1}. Since the DinoV2 model was recently introduced, the model performance has also been evaluated for other benchmark SDC datasets, such as HAM10000 and Dermnet, to test the robustness of the model on smaller datasets focusing on a relatively lesser number of classes, yet popular skin diseases. The authors believe the suggested study's practical impact is extremely valuable to doctors and the medical industry. The model's excellent best test accuracy of 96.48\% and F1-Score of 0.9728 (an improvement of approximately 10\% in accuracy and F1-Score over existing results) can aid these organizations in improving their ability to diagnose skin problems and offer patients more effective treatments. The interpretability of the results using the explainable AI (XAI) outputs, such as GradCAM and SHAP, obtained for test samples on the top-performing models used in this work can additionally guide dermatologists to perform clinical correlations and determine all the regions of occurrence. The dermatologists and the research community can utilize the results of this study to develop a mobile application for health organizations to swiftly and correctly identify skin problems, saving time and resources while increasing patient satisfaction with improved diagnosis and treatment. The major contributions of this work are:

\begin{itemize}
  \item Leveraging DinoV2 - a recently introduced transformer architecture alongside other transformer and convolutional neural networks to present state-of-the-art classification metrics on a geometrically augmented 31-class SDC dataset \cite{ref26}. This is the largest dataset known for the task and could provide accurate and quick diagnoses for a diverse set of skin ailments.
  \item Perform a comparative analysis on ConvNeXt - a benchmark CNN architecture for popular vision tasks, and all variants of three transformer architectures - ViT, Swin Transformers and DinoV2 on the augmented and unaugmented datasets to make a comprehensive choice of architecture. The training loss and accuracy curves are analyzed in depth to gain insights into the convergence of these models. 
  \item Evaluate the robustness of the proposed methodology by validating the performance on two smaller benchmark datasets: the HAM10000 and Dermnet datasets, with fewer samples and popular skin diseases
  \item Including the XAI results - SHAP and GradCAM, that would explain to dermatologists and clinicians why a particular disease is mapped to its corresponding label. This would help demystify the black-box nature of AI algorithms and assist dermatologists with the accurate and early diagnosis of skin diseases. The heatmaps or correlation plots also aid in determining the exact regions of infection, possibly giving more insights on severity to develop efficient treatment plans.
\end{itemize}

The manuscript is structured as follows: Section \ref{sec2} contains the related works done in the literature and the relevant gaps discovered and addressed. Section \ref{sec3} outlines the suggested technique, data curation, and experimental setup, whereas Section \ref{sec4} discusses the outcomes and models for the actual dataset, the explanation for the outputs for selected samples using XAI frameworks, and the outputs of the best-performing transformer architectures for the smaller datasets to test robustness. Section \ref{sec5} elucidates the limitations of the work and the future scope of improvement to accurately automate the SDC task.
Section \ref{sec6} concludes the work by summarizing it and describes the advantages that could be leveraged by medical professionals. Finally, sources utilized to identify literature are included in the final section of the manuscript.

\section{Related Works} \label{sec2}

The epidermis shields internal organs, which can get scarred or damaged due to infections or other factors such as worsening pollution and unhealthy diet. People commonly ignore the warning indications of a skin condition, and most current procedures for detecting and treating skin diseases rely on biopsies performed by a clinician. Since SDCs might be difficult to diagnose in a clinical context, the frequency of skin disorders has been growing, demanding quick and accurate detection \cite{ref47}. With the introduction of large-scale datasets such as ISIC 2018, \cite{ref36} HAM 10000 \cite{ref37} and Dermnet\cite{ref48}, several works in literature utilize deep learning models that can capture accurate features for feature classification with convolution and transformers. A proper diagnosis, assisted by these model predictions, can aid in the recovery from such ailments. 

Karthik et al. \cite{ref16} developed Eff2Net, a CNN that employs a channel attention block called ECA rather than the typical module to identify skin diseases. The model was evaluated on four diseases to obtain an accuracy of 84.70\%. Hossen et al. \cite{ref17} built a unique dataset of four dermatological diseases and compared a novel CNN with previous benchmark techniques. Image augmentation was also used to increase the size of the database and the model's scope. The model demonstrated good accuracy accuracy for the diseases. The combination of CNN-based SDC and a federated learning methodology provides an efficient way to classify skin diseases while protecting data. This motivated us to determine if augmentation additionally boosts results for the main dataset.

Andre Esteva et al. \cite{ref31} fine-tuned all the layers of InceptionV3 on a composite dataset to report a 72.1\% accuracy on the HAM10000 dataset. Kshirsagar et al. \cite{ref20} created a cutting-edge solution identifying skin problems with LSTM and MobileNetV2. The main goal of this research was to perform SDC correctly and determine if a hybrid technique can aid in preventing people. Though Saket S. Chaturvedi et al.  \cite{ref28} initially attempted to classify the HAM 10000 dataset using the ResNet101 backbone for feature extraction, they yielded better results, with an accuracy of 91.47\%. An improvisation was suggested by Anand et al.  \cite{ref19}, who suggested a pre-trained Xception model with transfer learning capability. The model was trained and tested on the HAM10000 dataset, classifying skin disorders with an accuracy of 96.40\%. With an accuracy of 99\%, the suggested model did exceptionally well in diagnosing Benign Keratosis. This strategy can help people and clinicians determine if medical intervention is required. Nevertheless, the authors of  \cite{ref27} proposed a fine-tuned Xception architecture to get high accuracy and an F1-score of 96\% on the 7-class MNIST HAM 10000 dataset, with data augmentation applied to prevent the class imbalance problem prevalent in the dataset to boost the results.

Hameed et al.  \cite{ref25} proposed an intelligent diagnostic technique for a more attractive cutaneous lesions class. The proposed approach was realized through hybrid techniques: error-correcting CNN and outcome codes based on a usable support vector machine. The study makes use of 9,144 images acquired from public sources. AlexNet, a CNN-approved approach, was used to extract the feature. Filali et al. \cite{ref29} used the PH2 dataset to detect melanoma using pre-trained and trained-from-scratch CNN models. They also applied preprocessing on the input image fed to the CNN using the Otsu algorithm to report an accuracy of 87.8\%. A similar study was carried out by Ly et al. \cite{ref30} with a model trained from scratch with a balanced PHDB dataset for classifying malignant skin cancer, with a reported accuracy of 86\% even without a publicly available HAM 10000 during their experimentation. There are works using ResNet \cite{ref45} and ResUNet \cite{ref46}, which get satisfactory results for the SDC task.

Some studies utilize private and custom datasets for SDC in their works. Velasco et al. \cite{ref24} introduced a model utilizing MobileNet for finding skin lesions with accuracy enhanced by using novel sampling strategies and preprocessing of input data. It was 84.28\% accurate using basic sampling methods. The accuracy was 93.6\%, with a skewed dataset and typical input record preparation. When oversampling in the dataset was found, the model's accuracy climbed to 91.8\%. Voggu and Rao \cite{ref22} suggested research in which three separate skin diseases would be detected using a novel approach. In this methodology, images of the skin are first preprocessed using filtering and alteration to reduce noise and undesired heredity. 

Since it was required to create automated methods for boosting analysis accuracy for various skin types and psoriasis symptoms, deep neural algorithms have been used to automatically detect skin problems. Bhavani et al. \cite{ref23} suggested a method for identifying various skin problems. Three examples, Mobile Net, Inception V3, and ResNet, are trained on a collection of machine learning features, notably logistic regression. Integrating the three CNNs in a hybrid architecture can result in excellent performance, though it reduces the evaluation's space and time complexity. The authors of \cite{ref26} were among the pioneers to do diverse work on a combined 31-class dataset, obtained by merging the non-overlapping and high sample quantity classes of two SDC datasets: Atlas Dermatology and ISIC 2018, containing 26 and 8 classes, respectively. The authors claim that the class count in the combined dataset is much higher than the benchmark datasets proposed in the literature. The results show that the EfficientNetB2 model performed the best with an 87.15\% accuracy for 31 classes of the augmented dataset.

Transformers have shown to be quite adept at handling complicated visual data. Their superior performance over CNNs in various visual tasks has been the driving force behind this revolution. They have become a potent substitute, processing picture patches through self-attentional processes. There has been a great deal of study towards improving transformer topologies due to their effectiveness in tasks like image classification, including skin diseases, as evident from the literature. Cai et al. \cite{ref14} demonstrated a multimodal Transformer for categorizing skin disorders. The architecture comprises dual encoders for pictures and metadata and a decoder for fusing the multimodal data. The proposed network employs a Vision Transformer model to extract deep features from pictures and incorporate metadata that serves as soft classification labels. In the decoder, the attention mechanism aids in the fusion of image and metadata characteristics. It was found to perform with an accuracy of 93.81\%, an improvement over state-of-the-art methods by 1\% on the ISIC dataset, making it a viable method for identifying skin diseases.

Aladhadh, Suliman, et al. \cite{ref35} were among the first to suggest Medical-VIT for Skin disease classification. Mild geometric and brightness-contrast-based augmentations helped their model fetch a test accuracy of 96.14\%. However, inspired by the work of \cite{ref33}, LesionAid \cite{ref15}, a novel multiclass prediction framework that classifies skin lesions based on ViT-GAN used GAN was used as an up-sampling algorithm to extract the genuine representation of the data from the raw images and synthesize new images to tackle the class imbalance problem. A model fine-tuned on such a synthetically up-sampled task yielded an immaculate validation accuracy of 97.4\% for classifying the HAM 10000 dataset using Vision transformers. The results were also closely followed by the one trained on Swin Transformers and its variants for the ISIS 2018 dataset by Selen Ayas \cite{ref34} to get an accuracy of 97.2\% using a weighted CCE loss in the Large22K model. 

\begin{figure*}[h]
\centering
\includegraphics[width=\textwidth]{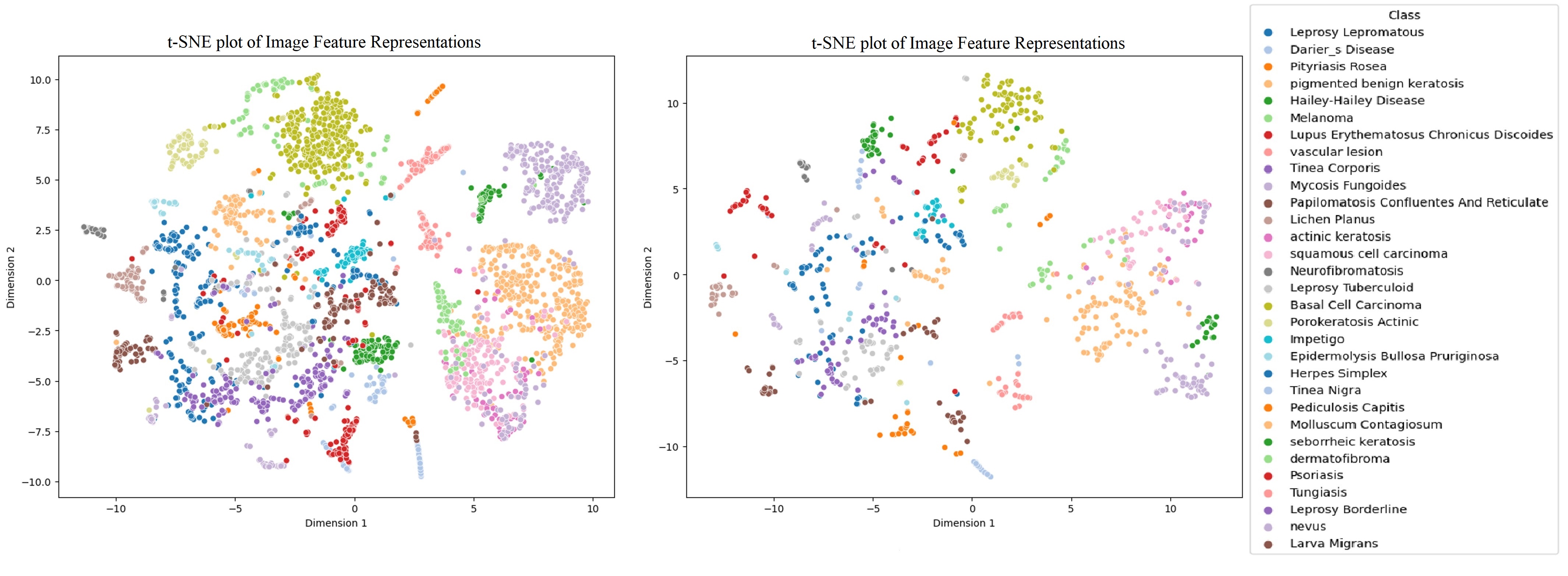}
\caption{t-SNE plot of the train (left) and test (right) data.}
\label{fig2}
\end{figure*}

Despite the improvement in the results of transformers on benchmark datasets and a few works using XAI to prove their efficiency, the models have been trained on smaller benchmark datasets to perform SDC. These datasets do not capture all regions in which diseases occur in the human body and different geographical areas of occurrence of these diseases and focus only on prominently occurring diseases. This may lead to diagnosing a rare disease as a popularly known disease that exhibits the same visible symptoms. With the growing number of skin disease cases belonging to a diverse category of infections, it is quintessential to accurately classify a much larger number of diseases containing more samples per class with a single transformer model. To the best of our knowledge, no study in the literature has used a complex transformer architecture like DinoV2 for a dermatology task. Thus, this work utilizes state-of-the-art transformers and performs transfer learning to improve prediction accuracy for a diverse 31-class dataset to improve the quality of diagnosis and prognosis of dermatological diseases. These results are compared with the benchmark results produced by CNN architectures for the dataset. the robustness of the proposed methodology is also tested by fine-tuning the model on other smaller datasets focusing on prominent dermatological problems.

In addition to experiments with state-of-the-art models, the black-box nature of the trained models is unravelled with the help of GradCAM and SHAP—two XAI frameworks that help dermatologists, doctors, and medical experts understand and visualize the regions of the image prioritized by each transformer to automate the diagnosis. This would assist them in diagnosing the disease more accurately and assist dermatologists with additional information like regions of occurrence that could be neglected because of human error. Addtionally, getting information on severity using heatmaps and extent and rate of spread can aid in administering treatment after cross-validating with patient clinical records.

\section{Methodology} \label{sec3}

\subsection{Dataset Description}\label{sec3.1}

Abdul Rafay and Waqar Hussain \cite{ref26} initially curated the dataset by combining the majority classes (categories with more than 80 samples) of the Atlas Dermatology and ISIC 2018 datasets, containing 3,399 and 561 images, respectively, to obtain a total of 4,910 samples. The dataset was split into an 80:20 train-test split. In our study, the train data was further split into a 90:10 split, resulting in an overall train-validation-test split of 72:8:20.

There were 561 different skin conditions listed in Atlas Dermatology, some of which lacked inadequate data to train and construct a deep model due to the scarcity of data. Even yet, just 9 to 10 samples were available for several classes. As a result, a threshold was established to curate the dataset manually, collecting data from classes with at least 80 examples. The dataset had 24 classes containing 3,399 samples after the filtration process. The second source, ISIC 2018, listed nine types of skin ailments. However, two of these nine classifications previously existed in the Atlas Dermatology dataset. Following filtration results, the two classes were omitted from the nine before the merger. The data from both sources was combined into a single dataset, and the resultant dataset had 31 classes and 4,910 samples in total. 

\begin{figure}[h]
\centering
\includegraphics[width=\linewidth]{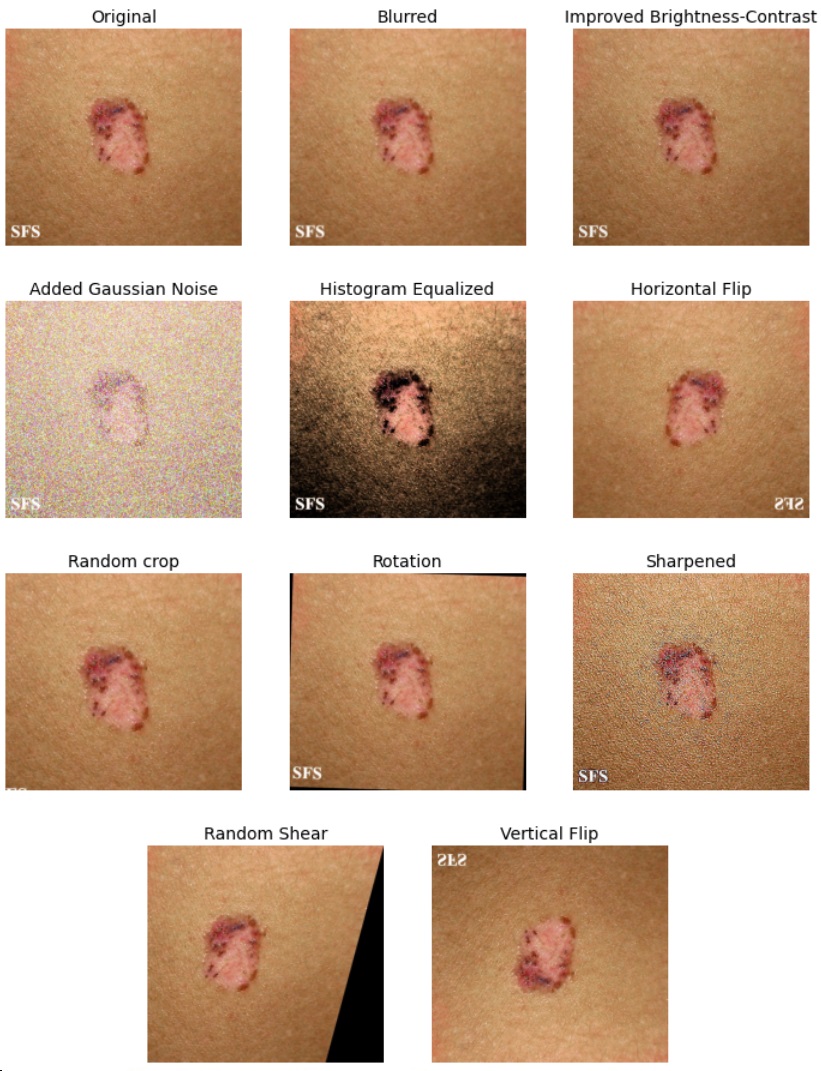}
\caption{Geometric augmentations used to upsample the dataset.}
\label{fig_new}
\end{figure}

\begin{figure*}[h]
\centering
\includegraphics[width=\textwidth]{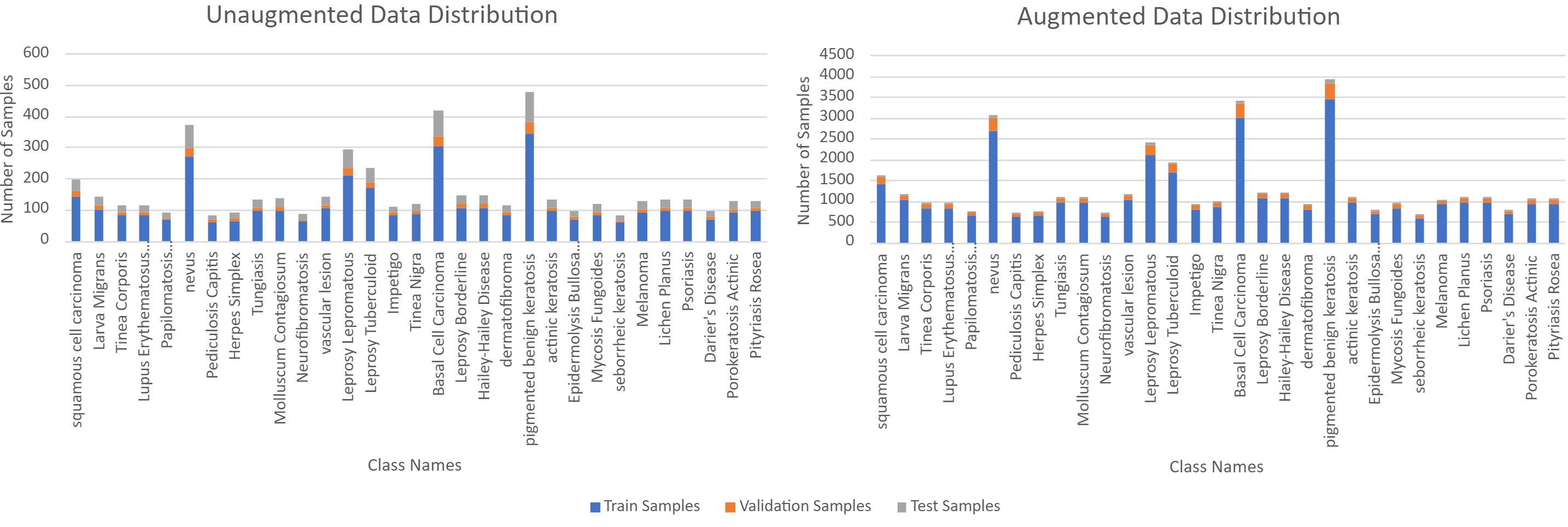}
\caption{Train-Validation-Test data distribution for the unaugmented/raw and augmented datasets.}
\label{fig3}
\end{figure*}

However, the distributions followed by these data samples are slightly different, as evident from the train and test T-Stochastic Nearest Embedding (t-SNE) plots in Figure \ref{fig2}. The train distribution has samples of the same class that are more cluttered together, indicating that training a model to classify samples belonging to different classes would not be difficult. However, the test dataset t-SNE plot shows samples more distributed in space, indicating a difficult linear separability. The dataset can be oversampled using augmentations to make the model more robust. Thus, ablation experiments with ten different types of augmentations: Vertical and Horizontal Flipping, Random Shear, Sharpening, Random Rotation, Center Crop, Brightness, and Contrast variation, Histogram Equalization, Gaussian Noise, and blurring were used to up-sample the training dataset to oversample the train data, to determine if the attempt improves the overall test accuracy. The appearance of samples post data augmentation for a randomly chosen sample belonging to the class "Basal Cell Carcinoma" is shown in Figure \ref{fig_new}. The insights into the number of samples in different partitions of the data are mentioned in Table \ref{table1}, and the data distribution for each class for the split is mentioned in Figure \ref{fig3}.

\begin{table}[ht]
    \centering
    \caption{Sample distribution of the main dataset.}
    \label{table1}
    \begin{tabular}{|c|c|c|c|c|}
        \hline
        & \textbf{Train} & \textbf{Validation} & \textbf{Test} & \textbf{Total}  \\
        \hline
        \textbf{Raw Data} & 3,524 & 392 & 994 & 4,910 \\
        \hline
        \textbf{Augmented Data} & 35,240 & 3,920 & 994 & 49,100 \\
        \hline
    \end{tabular}
\end{table}

Apart from this dataset, two smaller benchmark datasets have also been considered for analyzing the robustness of transformer architectures for the SDC task that contain images of popular skin diseases. The HAM10000 dataset, which comprises image samples covering important diagnostic categories like actinic keratoses and other pigmented lesions, is a large collection of multi-source dermatoscopic images of common pigmented skin lesions, providing valuable resources for research and classification purposes. It contains 10,015 images belonging to 7 classes. Another dataset called Dermnet is a collection of images used for the localization and classification of various skin diseases. A 23-class dataset with 19,500 images is maintained by a diverse group of dermatologists and contains images representing different skin conditions for research and diagnostic purposes. Table \ref{table2} contains the number of samples present in the additional datasets that are benchmarked in this work.

\begin{table}[ht]
    \centering
    \caption{Sample distribution of the additional datasets.}
    \label{table2}
    \begin{tabular}{|c|c|c|c|c|}
        \hline
        \textbf{Dataset} & \textbf{Train} & \textbf{Validation} & \textbf{Test} & \textbf{Total}  \\
        \hline
        HAM10000 & 7,211 & 801 & 2,003 & 10,015 \\
        \hline
        Dermnet & 13,950 & 1,550 & 4,000 & 19,500 \\
        \hline
    \end{tabular}
\end{table}

\subsection{Transformer Networks used} \label{sec3.2}

Transformers have outperformed classic CNNs in image classification, object identification, and other computer vision tasks, opening the way for integrating text and picture information in multimodal applications. As they continue to impact the computer vision environment, research focuses on refining their design, scaling them to bigger datasets, and examining their potential for tackling various visual comprehension difficulties, including essential biomedical applications. What makes the proposed study unusual is no previous research has been undertaken utilizing transformers such as DinoV2 on a dermatology task, to our knowledge. Moreover, this dataset helped us comprehensively analyze SDC with other popular transformers on the biggest SDC dataset. Thus, in addition to the benchmark convolution networks that have been used in the literature, the following transformers were trained on the three datasets to validate their performance on the test data split and use the metrics for the comparative analysis.

\subsubsection{Vision Transformers} \label{sec3.2.1}
Because of their exceptional performance and scalability, ViTs \cite{ref38} have received much interest in image classification. Unlike typical CNNs, ViTs employ a transformer architecture initially built for natural language processing workloads. ViTs divide an image into non-overlapping patches and embed them linearly into a series of tokens, which are subsequently processed by transformer layers. The equation of the output computed by the multi-head self-attention block on the embeddings is given in equation \ref{eqnF1} and \ref{eqnF2}. It enables ViTs to record long-term relationships and contextual information over the whole image, making them helpful in dealing with complicated visual patterns.

The architecture has demonstrated the ability to handle pictures of changing sizes without requiring substantial architectural adjustments. ViT models pre-trained on large-scale datasets have demonstrated high transfer learning capabilities, allowing for fine-tuning on smaller datasets for specialized image classification tasks. On the other hand, they may be computationally costly and require a large quantity of training data to work well. Nonetheless, they are a promising trend in image classification and are constantly improving, with researchers investigating different architectural enhancements and training strategies to increase their performance.

\begin{equation}
\label{eqnF1}
\text{MHSA}_{Q, K, V} = \text{Concat}(\text{head}_1, \ldots, \text{head}_h)W^O
\end{equation}

\begin{equation}
\label{eqnF2}
\text{head}_i =\text{softmax}\left(\frac{Q_iK_i^T}{\sqrt{d_k}}\right)V_i
\end{equation}

\subsubsection{Swin Transformers} \label{sec3.2.2}

Swin Transformers \cite{ref39} is yet another novel way of image categorization that has shown to be quite effective. Swin Transformers overcome some of the limitations of classic CNNs and ViTs by employing a hierarchical design that effectively gathers local and global information. Swin Transformers, like ViTs, divide the picture into non-overlapping patches, but unlike ViTs, Swin Transformers employ a hierarchical design with many stages. Each stage comprises a series of transformer layers that analyze data at various spatial resolutions.

Swin Transformers' computational efficiency is one of its primary advantages. They lower total computing costs with a linear complexity compared to ViT's quadratic complexity while retaining comparable performance by processing information hierarchically. As a result, they are more suitable for real applications requiring minimal processing resources. The model has demonstrated outstanding performance on various image classification standards and remains an active field of study in medical image classification due to its ability to balance efficiency with efficacy.

\begin{figure*}[h]
\centering
\includegraphics[width=\textwidth]{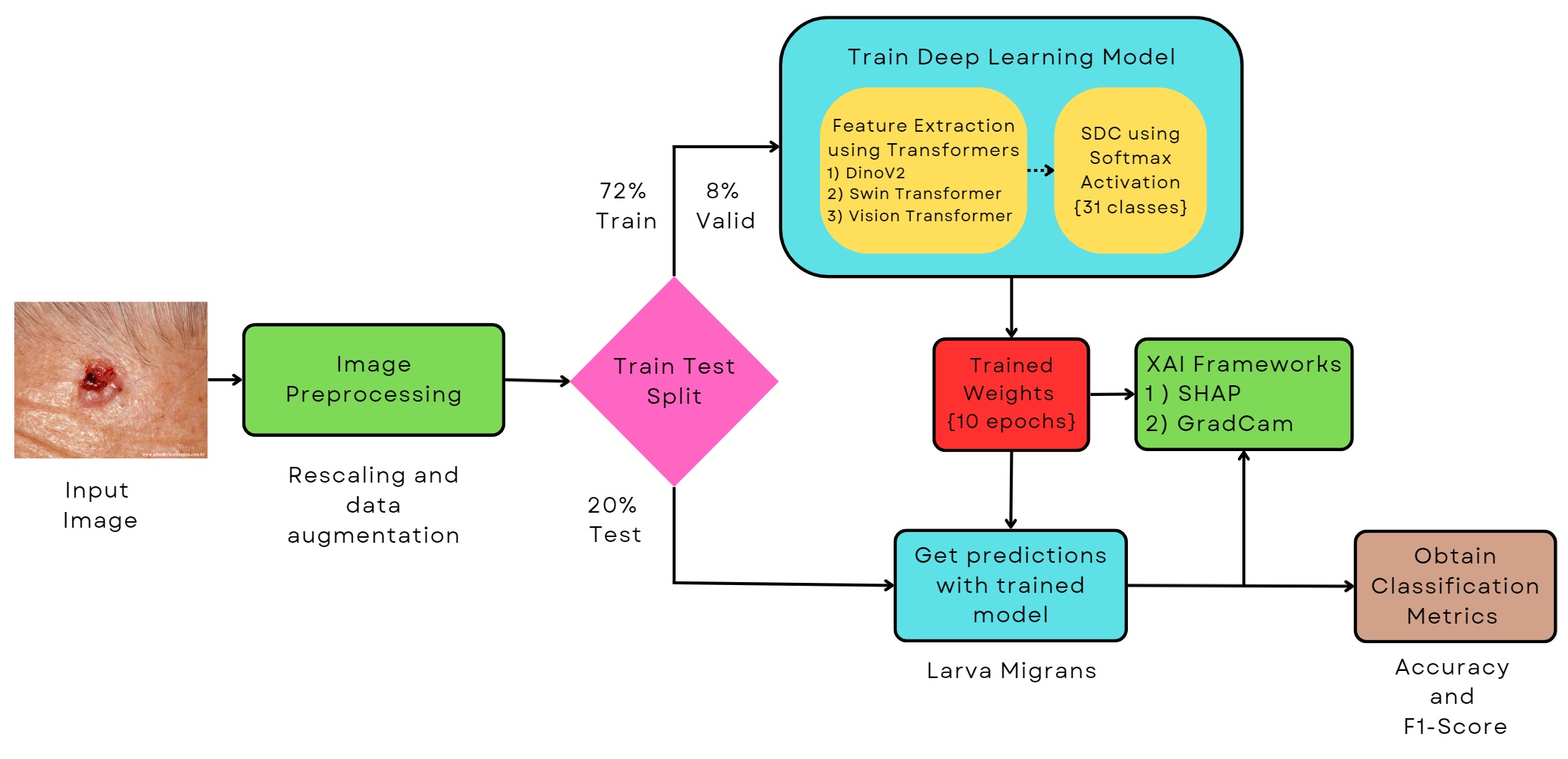}
\caption{Overall methodology proposed in this work}
\label{fig4}
\end{figure*}

\subsubsection{DinoV2} \label{sec3.2.3}

The self-DIstillation with NO labels (DINO) \cite{ref40} is a sophisticated self-supervised learning approach for training models that improves computer vision by reliably detecting specific objects inside pictures and video frames. Many academics and organizations have concentrated their efforts on self-supervision learning (SSL) models in recent years, generating labels using a semi-automatic method that entails watching a labelled dataset and estimating part of the data from that batch based on the characteristics. Some SSL systems circumvent these issues by employing DINO, which employs SSL and knowledge distillation methods. It enables extraordinary features to develop, such as robust object component recognition and robust semantic and low-level picture understanding.

DINOv2 addresses the issue of training larger models with more data by enhancing stability through regularization approaches inspired by the similarity search and classification literature and incorporating efficient PyTorch 2 and xFormers techniques. It leads to quicker, more memory-efficient training with the potential for data, model size, and hardware scaling. In addition to the approaches, the researchers also applied parameters such as the iBOT Masked Image Modeling (MIM) loss term, the curriculum learning strategy to train the models in a meaningful order from low to high-resolution images, softmax normalization, KoLeo regularizers (which improve the nearest-neighbour search task), and the L2-norm for normalizing the embeddings are some of the strategies DINOv2 adopted to improve their results.

\subsection{XAI for explainability} \label{sec3.3}

Explainable Artificial Intelligence, or XAI, is an important AI research and development topic as it tries to improve the transparency and interpretability of AI systems, allowing people to comprehend their decision-making processes. XAI solves the black box issue that frequently afflicts complicated machine learning models such as deep neural networks. XAI increases trust and responsibility by offering insights into why AI systems make certain predictions or conclusions. Still, it also helps users uncover and minimize biases, mistakes, and unexpected behaviours in AI applications. XAI employs various approaches and procedures, from visualization to feature attribution, aiming to make AI systems more interpretable and accessible to professionals and non-experts.

GradCAM, a computer vision algorithm, stands for Gradient-weighted Class Activation Mapping. It creates heatmaps emphasizing parts of an input picture that contribute the most to a deep neural network's classification judgment. This graphic explanation explains which components of a picture were important in the model's decision-making process. SHAP is a larger technique that may be used in a wide range of machine learning models, including some unrelated to computer vision. SHAP values are based on cooperative game theory and give a mechanism to ascribe the contribution of each characteristic to a certain prediction or result. This method thoroughly explains how specific input features impact model output, making it useful for model interpretation and feature engineering.

While GradCAM is particularly beneficial for visualizing deep neural network judgments in image-related tasks, SHAP offers a more adaptable technique that can be applied to various machine learning models and is particularly good for determining feature significance. Both strategies contribute to the larger subject of XAI by improving AI system transparency and interpretability.

\subsection{Experimental Setup} \label{sec3.4}

Twenty experiments - Ten different architectures belonging to four backbones, and two types of SDC datasets (with and without data augmentation) were done in this study. The experiments were done in a system with an Nvidia RTX A6000 GPU with 48GB vRAM, a Ryzen Threadriper Pro CPU with 120GB RAM, and 24 cores. The proposed pipeline to carry out the study done in this work is shown in Figure \ref{fig4}. 

The models were trained for 10 epochs on all backbones of the transformers used and were fed in batches of 64 to these models. The models were trained on both the unaugmented and augmented versions of the dataset to determine if augmentation leads to overfitting on train data, as deciphered from the train and test T-SNE plots shown in Figure \ref{fig1}. The geometric augmentations, as explained in the methodology, were meticulously chosen to enhance the diversity and robustness of our dataset, aiming to expose models to various data distributions and real-world variations.

The PyTorch framework and the weights from the Huggingface library were used to code and perform transfer learning using pre-trained ImageNet1k weights. Categorical cross entropy (CCE) loss was used to calculate the classification error during the training backpropagation process (the equation to calculate the loss is given in Equation \ref{eqnL}), and Adam was used as the optimizer for faster training. The optimal learning rate during gradient descent was calculated using the lr\_find() function, which divides the data into batches and considers choices from the learning rate yielding the least loss. 

\begin{equation}
\label{eqnL}
\text{Loss}_{\text{CCE}} = -\sum_{i=1}^{N} \sum_{j=1}^{K} y_{ij} \log(p_{ij})
\end{equation}

All these models are evaluated on the test split of the dataset with popular classification metrics such as accuracy, precision, recall and F1-score. Equations \ref{eqn1}-\ref{eqn4} denote the formulas for calculating the classification metrics. The results were also explained visually using XAI Tools such as GradCAM and SHAP to get more insights into the features captured by the model to diagnose a disease.

\begin{equation}
\label{eqn1}
\text{Accuracy} = \frac{\text{Number of Correct Predictions}}{\text{Total Number of Predictions}}
\end{equation}

\begin{equation}
\label{eqn2}
\text{Precision} = \frac{\text{True Positives}}{\text{True Positives} + \text{False Positives}}
\end{equation}

\begin{equation}
\label{eqn3}
\text{Recall} = \frac{\text{True Positives}}{\text{True Positives} + \text{False Negatives}}
\end{equation}

\begin{equation}
\label{eqn4}
\text{F1 Score} = \frac{2 \cdot \text{Precision} \cdot \text{Recall}}{\text{Precision} + \text{Recall}}
\end{equation}

\section{Results and Discussion} \label{sec4}

\begin{table*}[h]
\centering
\caption{Classification metrics for the architectures trained on the unaugmented and augmented versions of the combined SDC dataset.}
\label{tab2}
\begin{tabular}{|c|c|cccc|cccc|}
\hline
\textbf{Model} &
  \textbf{Parameters} &
  \multicolumn{4}{c|}{\textbf{Augmented Data}} &
  \multicolumn{4}{c|}{\textbf{Unugmented Data}} \\ \hline
\multicolumn{1}{|l|}{} &
  \multicolumn{1}{l|}{} &
  \multicolumn{1}{c}{\textbf{Accuracy}} &
  \multicolumn{1}{c}{\textbf{Precision}} &
  \multicolumn{1}{c}{\textbf{Recall}} &
  \textbf{F1 score} &
  \multicolumn{1}{c}{\textbf{Accuracy}} &
  \multicolumn{1}{c}{\textbf{Precision}} &
  \multicolumn{1}{c}{\textbf{Recall}} &
  \textbf{F1 score} \\ \hline
  \textbf{ConvNeXt-B} &
  87,598,239 &
  \multicolumn{1}{c}{83.10} &
  \multicolumn{1}{c}{85.04} &
  \multicolumn{1}{c}{83.58} &
  83.98 &
  \multicolumn{1}{c}{31.18} &
  \multicolumn{1}{c}{17.53} &
  \multicolumn{1}{c}{15.92} &
  12.19 \\ \hline
\textbf{Swin-T} &
  27,543,193 &
  \multicolumn{1}{c}{39.43} &
  \multicolumn{1}{c}{42.96} &
  \multicolumn{1}{c}{28.81} &
  29.01 &
  \multicolumn{1}{c}{36.01} &
  \multicolumn{1}{c}{49.71} &
  \multicolumn{1}{c}{26.06} &
  26.55 \\
\textbf{Swin-S} &
  48,861,097 &
  \multicolumn{1}{c}{84.41} &
  \multicolumn{1}{c}{86.14} &
  \multicolumn{1}{c}{86.04} &
  85.64 &
  \multicolumn{1}{c}{40.44} &
  \multicolumn{1}{c}{42.8} &
  \multicolumn{1}{c}{29.29} &
  29.33 \\
\textbf{Swin-B} &
  86,774,999 &
  \multicolumn{1}{c}{90.44} &
  \multicolumn{1}{c}{92.16} &
  \multicolumn{1}{c}{92.64} &
  92.31 &
  \multicolumn{1}{c}{93.26} &
  \multicolumn{1}{c}{94.88} &
  \multicolumn{1}{c}{95.15} &
  94.71 \\
\textbf{Swin-L} &
  195,043,123 &
  \multicolumn{1}{c}{88.93} &
  \multicolumn{1}{c}{91.14} &
  \multicolumn{1}{c}{90.95} &
  90.89 &
  \multicolumn{1}{c}{75.85} &
  \multicolumn{1}{c}{79.23} &
  \multicolumn{1}{c}{75.62} &
  76.55 \\ \hline
\textbf{ViT-B} &
  85,822,495 &
  \multicolumn{1}{c}{94.37} &
  \multicolumn{1}{c}{95.62} &
  \multicolumn{1}{c}{95.60} &
  95.51 &
  \multicolumn{1}{c}{92.35} &
  \multicolumn{1}{c}{93.67} &
  \multicolumn{1}{c}{93.70} &
  93.49 \\
\textbf{ViT-L} &
  303,333,407 &
  \multicolumn{1}{c}{88.63} &
  \multicolumn{1}{c}{90.67} &
  \multicolumn{1}{c}{90.55} &
  90.44 &
  \multicolumn{1}{c}{87.22} &
  \multicolumn{1}{c}{88.79} &
  \multicolumn{1}{c}{88.53} &
  88.39 \\ \hline

\textbf{DinoV2-S} &
  \textbf{22,080,415} &
  \multicolumn{1}{c}{87.62} &
  \multicolumn{1}{c}{89.85} &
  \multicolumn{1}{c}{89.24} &
  89.37 &
  \multicolumn{1}{c}{60.26} &
  \multicolumn{1}{c}{62.50} &
  \multicolumn{1}{c}{58.08} &
  58.52 \\
\textbf{DinoV2-B} &
  86,628,127 &
  \multicolumn{1}{c}{95.57} &
  \multicolumn{1}{c}{96.81} &
  \multicolumn{1}{c}{96.72} &
  96.71 &
  \multicolumn{1}{c}{\textbf{96.48}} &
  \multicolumn{1}{c}{\textbf{97.55}} &
  \multicolumn{1}{c}{\textbf{97.10}} &
  \textbf{97.27} \\
\textbf{DinoV2-L} &
  304,432,159 &
  \multicolumn{1}{c}{90.44} &
  \multicolumn{1}{c}{92.64} &
  \multicolumn{1}{c}{92.22} &
  92.33 &
  \multicolumn{1}{c}{88.02} &
  \multicolumn{1}{c}{89.65} &
  \multicolumn{1}{c}{89.95} &
  89.60 \\ \hline
\end{tabular}
\end{table*}

The experiments done in this work leverage three transformers: Vision Transformers, Swin Transformers, and DinoV2. The ConvNeXt architecture, a benchmark in convolution-based feature extraction for image classification tasks has also been trained and validated for the main 31-class dataset. Other convolution architectures that have been adopted as backbones for feature extraction in the literature have also been used to extend the comparative analysis. The models are interpreted using XAI frameworks to assist dermatology and unravel the black-box nature of deep learning.

\begin{figure*}[h]
\centering
\includegraphics[width=\textwidth]{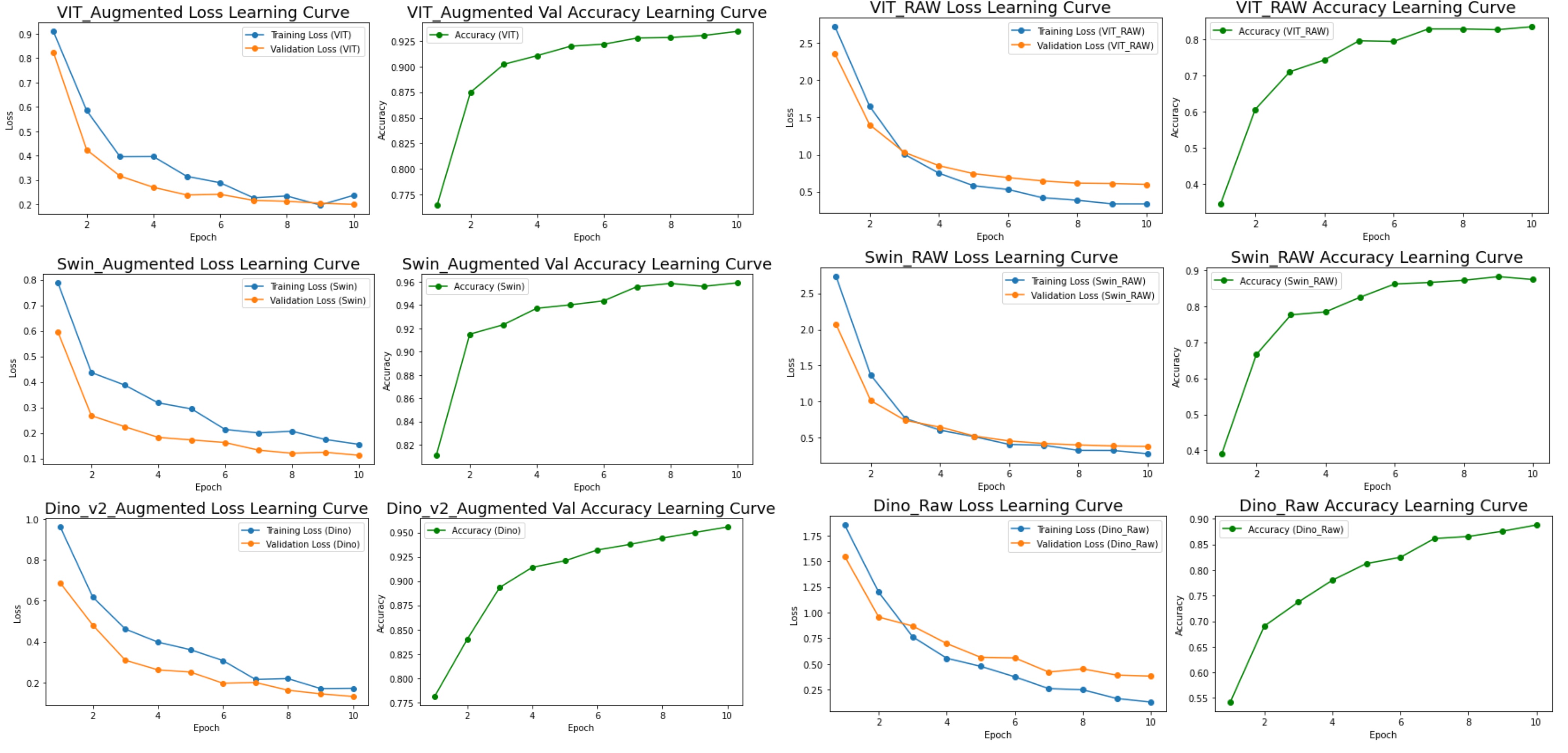}
\caption{Epoch vs. Loss and Accuracy curves for all trained models}
\label{fig5}
\end{figure*}

Additionally, to ascertain if the best-performing transformer model on the 31-class dataset is robust, the model is also trained on two smaller datasets containing more samples of prominent diseases, and the metrics are compared with those of other benchmark models proposed in the literature. 

\subsection{Results on the combined SDC dataset.}\label{sec4.1}

Table \ref{tab2} shows the classification metrics obtained on the augmented and raw datasets using all the official releases of the three different transformer models used in this work. From the comparative analysis of transformer-based architectures alongside the convolutional-based architecture ConvNeXt, the results reveal several key insights from the perspective of deep learning for medical image analysis. The benchmark convolution architecture ConvNeXt-B demonstrates comparatively lower performance with just 31.18\% accuracy, particularly on unaugmented data, where it struggles to generalize effectively. This suggests potential limitations in ConvNeXt-B's ability to adapt to diverse datasets without augmentation. This could be due to the fact that convolution extracts local features with the assistance of a kernel, limiting its scope to that region alone for feature extraction in a particular layer, but the attention mechanism ensures correlation computation between all patches of the image feature map in a layer.

ViT improves these results because it considers the relationship between the image's different fixed patch embeddings. Yet, Swin Tranformers perform better than ViT due to an improved sliding kernel attention mechanism.

Additionally, from all the classification metrics for the test results of the experimental results, it is clear that the general performance trend of the models trained on the augmented SDC dataset is better than those trained on the unaugmented data, suggesting that the train and test distributions are indeed not as different as deciphered from the t-SNE plots. An improvement in accuracy and a similar improvement in other metrics is noticeable for all backbones trained on augmented data, except for the Swin-B and DinoV2-B backbones, whose metrics deteriorate post-augmentation. Though DinoV2-B experiences a drop of 1\% in all metrics, the drop is not very significant for recall, denoting a lesser increase in the number of false negative predictions. Nevertheless, all the classification metrics of DinoV2-B consistently outperform other models across all metrics, showcasing its effectiveness in both augmented and unaugmented data scenarios. This suggests that the self-supervised pre-training method utilized in DinoV2-B yields superior results compared to the supervised pre-training approach adopted by other transformer models. The slight drop in performance metrics for DinoV2-B post-augmentation indicates a potential overfitting to the augmented training data, leading to a marginal decline in test results. This overfitting suggests that while augmentations can enhance model robustness, they might also introduce noise that affects generalization, particularly in SSL models like DinoV2-B. DinoV2's consistent performance superiority, even with fewer training samples, underscores its robustness and efficiency. The self-supervised pre-training method enables the model to learn more generalized features from the data, making it less reliant on large annotated datasets.. Also, the performance of Swin Transformers drops below ViT's post augmentation, with a mere 90.44\% accuracy, and all metrics drop by 2\% despite architectural dominance, indicating the importance of model size in achieving higher metrics.

On the other hand, one can also infer from the results of all backbones within a family that the smaller models might generalize better than larger ones, especially in cases where the dataset is diverse and representative of the target domain. Yet, larger models tend to have more parameters, making them more prone to overfitting, especially when the dataset is not large enough to fully exploit the model's capacity. For the combined unaugmented data considered for the study, the training data per class might be limited (just like it is seen for a few classes in \ref{fig3}), and overfitting is more prominent, leading to a decrease in performance for larger models. This is justified by the generic trend in results across each family of transformers, where due to the size of the combined dataset and the trainable parameters of the model, the classification metrics can be easily observed to increase from tiny up to the base models of all architectures, but a small drop in the performance of the large variant is observed.

The classification metrics, in general, are better for the augmented version of the dataset than the unaugmented version, suggesting that the augmentation strategies adapted to upsample the dataset are indeed helpful in helping the generalization of relevant features extracted by the architectures. The accuracy improvement is about 10\% in cases where classification metrics have improved in general. However, this is not the case for the Swin-B and DinoV2-B models due to overfitting. The overfitting nature of the model trained on augmented data can also be substantiated by the epoch vs. loss and accuracy curves for all the best-performing variants of the chosen transformer models shown in Figure \ref{fig5}. Firstly, the horizontal gap between the train and validation loss curves keeps fluctuating for the models trained with the augmented dataset. Still, a vast fluctuation is absent for the models trained on unaugmented data. This erratic fluctuation in loss curves for augmented data indicates that the models may struggle to generalize effectively, leading to increased overfitting. Upon closer examination of the loss curves on the y-axis, the values are consistently smaller for augmented data compared to the other models for the same epoch. However, the corresponding improvement in validation accuracy is not observable. 

Moreover, a distinct trend emerges when assessing the validation accuracy curves. Models trained with data augmentation tend to rapidly reach high accuracy levels, often within the first few epochs, before saturating. This is evidenced by the validation accuracy higher for training with data augmentation than the data due to the validation data being a subset of the train data and the model becoming well-trained on the train data samples. However, overfitting with an augmented dataset leads to overtraining, yielding lesser classification metrics for the test data. In contrast, models trained on the raw dataset exhibit a more gradual and steady increase in accuracy over time. This phenomenon can be attributed to the overfitting observed in augmented data, where the models essentially 'memorize' the training samples rather than learning generalized patterns. Though the models obtain almost the same quantity of false positives and false negatives, owing to a comparable precision, recall, and F1-score for their unaugmented counterparts, the lesser true positives and true negatives (as deciphered from the accuracy) make the model performance relatively poor. Thus, though augmentation strategies are helpful in general, it is not necessary for architectures like Swin transformers and DinoV2, as demonstrated by the results.

\begin{figure}[h]
\centering
\includegraphics[width=\linewidth]{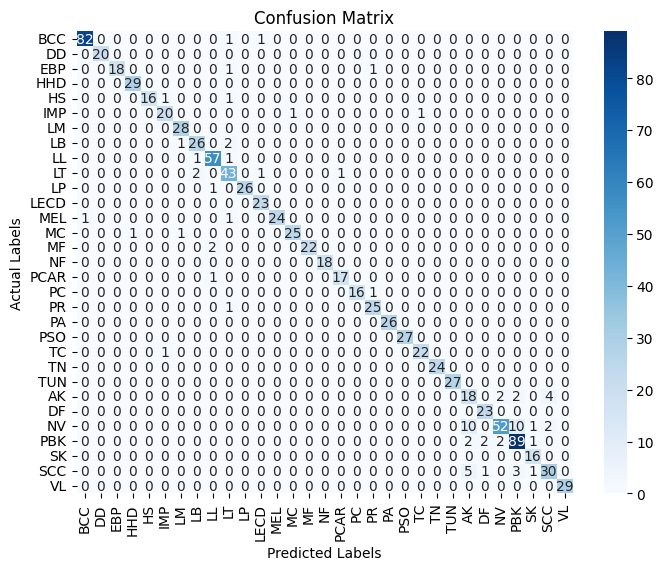}
\caption{Confusion matrix for the trained ViT-Base model on unaugmented data.}
\label{fig6}
\end{figure}

\begin{figure}[h]
\centering
\includegraphics[width=\linewidth]{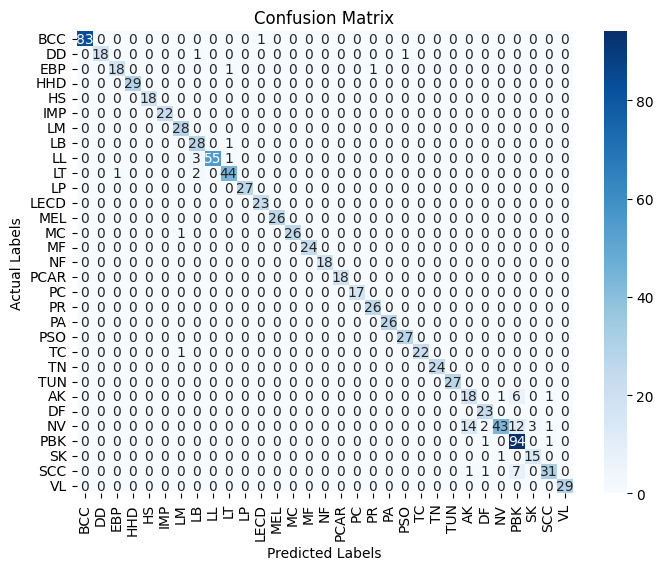}
\caption{Confusion matrix for the trained Swin-Base model on unaugmented data.}
\label{fig7}
\end{figure}

\begin{figure}[h]
\centering
\includegraphics[width=\linewidth]{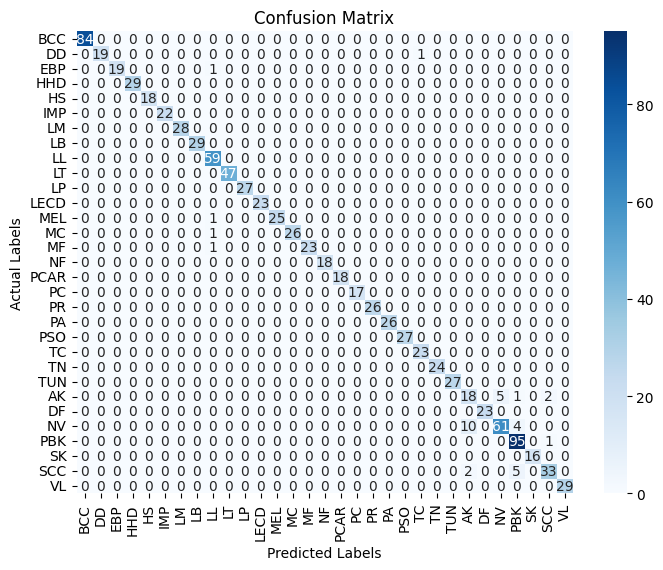}
\caption{Confusion matrix for the trained DinoV2-Base model on unaugmented data.}
\label{fig8}
\end{figure}

Since the best performing models were acquired with training on an unaugmented dataset, Figures \ref{fig6},\ref{fig7}, and \ref{fig8} show the confusion matrices obtained by the models ViT-B, Swin-B, and DinoV2-B, respectively, trained on the unaugmented data. While comparing the metrics of each model, the performance of Vision Transformers is the lowest compared to Swin Transformers and DinoV2. Though ViT can theoretically extract better feature maps than CNNs using the multi-head self-attention layer from the patch and position embeddings, due to which the model gets a test accuracy above 90\%, the model is outperformed by the sliding window self-attention blocks of the Swin Transformers. Moreover, a higher classification metric of 93.26\% accuracy and a 0.95 Macro-F1 score, which is approximately a 1\% improvement over the accuracy of ViT makes Swin a better architecture to perform the task. The model obtained a lesser number of false positive and false negative values. However, the standout performer in our experiments is DinoV2, a network pre-trained with semi-supervised approaches, which harnesses the Xformers framework to attain the best test accuracy of 96.48\% and the least outliers (less than 40 of the 944 samples). This substantial improvement in accuracy positions DinoV2-B as the most promising architecture among the tested models, surpassing both Swin Transformers and ViT for image classification tasks. Nevertheless, the model does have a few outliers (elements not along the diagonal) in the confusion matrix, suggesting room for improvement. 

Another standard inference from the confusion matrices of all three models is a significant number of samples (10 or higher) in the test set of the Nevus (N) class, being incorrectly predicted as the Actinic Keratosis (AK). This is because AK and nevus can sometimes be misdiagnosed due to overlapping clinical features. AK presents as scaly, rough patches, often on sun-exposed areas, while nevi (moles) are pigmented skin growths. However, certain types of nevi, such as dysplastic nevi, may exhibit features resembling AK, leading to misdiagnosis. Additionally, both conditions can arise from sun exposure, further complicating diagnosis. Furthermore, the differential diagnosis may be challenging, as flat pigmented lesions on sun-damaged skin, including nevi, can mimic actinic keratosis, which the transformer architectures can't easily decipher from the training dataset.

\begin{figure*}[h]
\centering
\includegraphics[width=\textwidth]{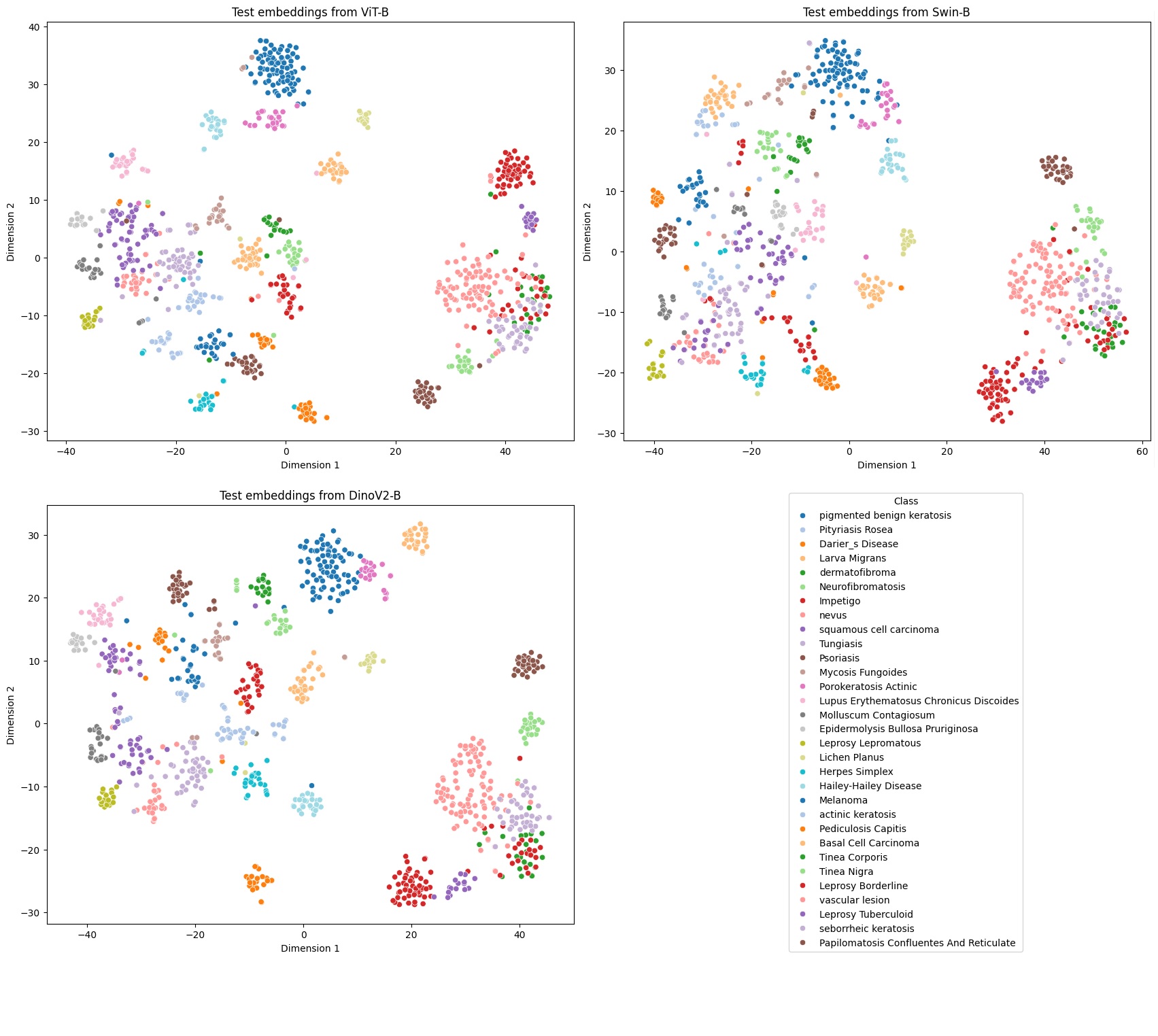}
\caption{T-SNE plot for the embeddings obtained from the fine-tuned transformer architectures on the unaugmented dataset.}
\label{fig9}
\end{figure*}

Figure \ref{fig9} shows the t-SNE plots for the feature mapvectors extracted from ViT-B, Swin-B, and DinoV2-B. In all models, the class clusters of the test embedding are closer to each other, demonstrating structural similarities for samples within the same class. However, owing to inter-class similarities, some embedding projections are distributed throughout the space and overlap with closely related classes, explaining why the task itself is primarily difficult even for robust transformer-based feature extractors. Nevertheless, a model such as DinoV2, which is robustly trained well on the combined dataset, performs better, only on a dataset close to its distribution, as evident from the points being closer to the corresponding cluster centres (demonstrating low intra-class variability) and the high classification metrics obtained by the model in this work.

\begin{figure}
    \centering
    \includegraphics[width=\linewidth,height=20cm]{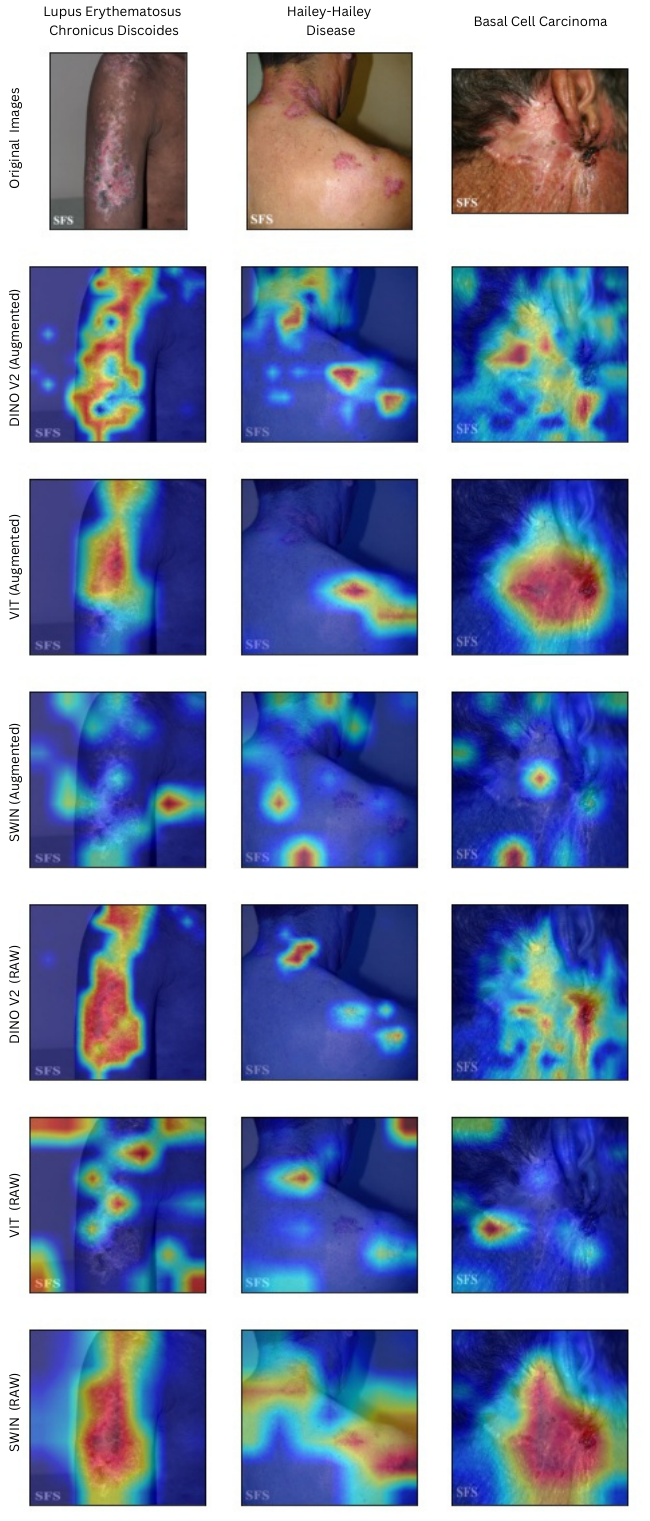}
    \caption{Gradcam plots for Augmented \& Raw Images}
    \label{fig10}
\end{figure}

\begin{figure}
    \centering
    \includegraphics[width=\linewidth,height=20cm]{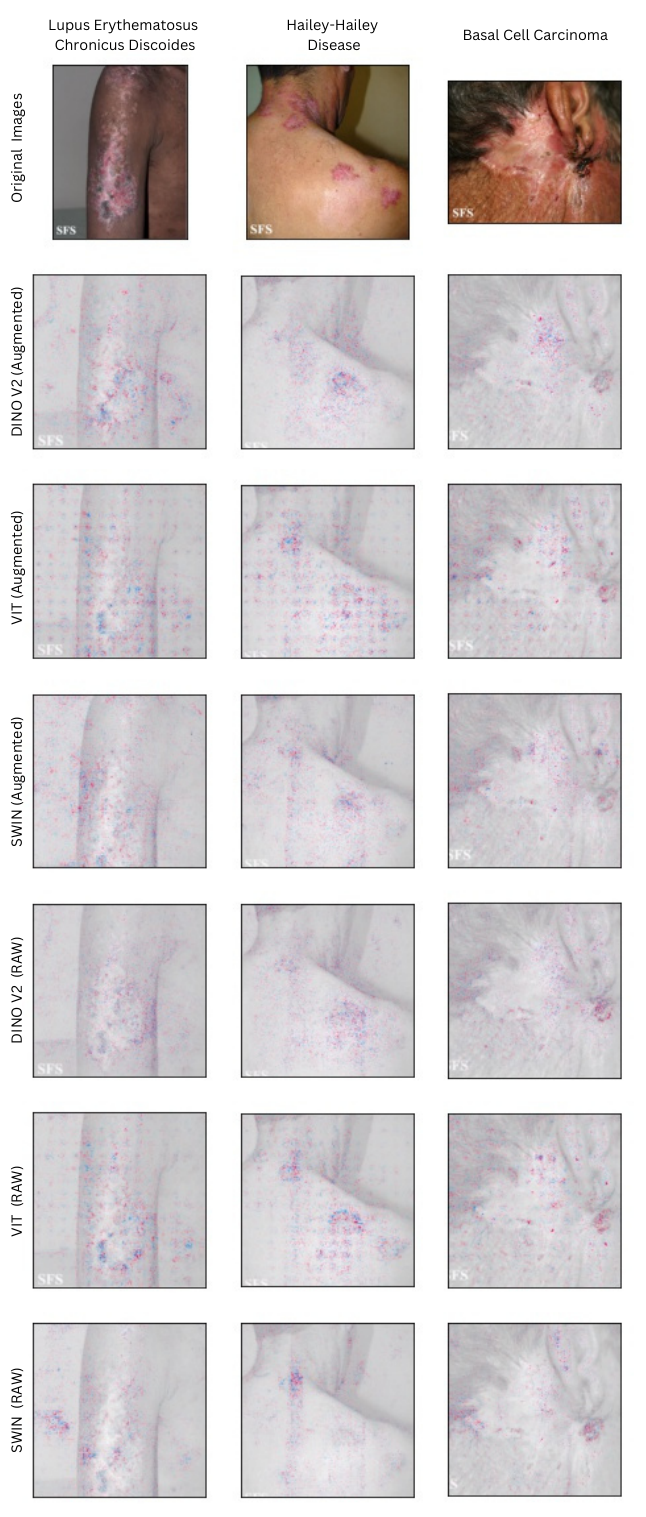}
    \caption{SHAP plots for Augmented \& Raw Images}
    \label{fig11}
\end{figure}

\begin{table*}[h]
\centering
\caption{Comparison of SDC Models on combined data}
\label{tab3}
\begin{tabular}{|c|c|c|c|c|c|c|c|c|c|c||}
\hline
\textbf{Authors} & \textbf{Year} & \textbf{Classes} & \textbf{Architecture} & \textbf{Accuracy}& \textbf{Precision} & \textbf{Recall}& \textbf{F1-Score}\\
\hline
A. Rafay and W. Hussain \cite{ref26}& 2023 & 31 & EfficentNetB2 & 0.8715 & 0.87& 0.87& 0.87 \\
\hline
\textbf{Proposed Work} &  & 31 & ViT-Base & 0.9235 & 0.9367 & 0.9370 & 0.9349\\
 &  & 31 & Swin-Base & 0.9326 & 0.9488 & 0.9516 & 0.9472\\
 &  & \textbf{31} & \textbf{DinoV2-Base} & \textbf{0.9648} & \textbf{0.9755} & \textbf{0.9711} & \textbf{0.9728}\\
\hline
\end{tabular}
\end{table*}

Table \ref{tab3} highlights the model performance and classification metrics of the experiments done with the transformer-based architectures alongside the state-of-the-art models trained on the dataset used in this work. The only work on this dataset was done by the group that introduced the dataset, and they adopted convolution-based architectures from a family of architectures such as EfficientNet, VGG, and ResNet. to conclude that EfficentNetB2 achieves the best classification accuracy. Nevertheless, all transformers used in our work perform better and yield better benchmark results, with DinoV2-B improving the accuracy by approximately 10\% to yield a 96.48\% accuracy compared to an existing 87.15\% accuracy. Thus, our study underscores the significance of transformer model architectures, pre-training strategy, and data augmentation in determining the performance of deep learning models for image classification tasks, with DinoV2-B emerging as the top-performing transformer architecture in this comparative analysis. This potentially paves the way for classifying many such medical image datasets in the future.

\subsection{Explanability using XAI frameworks}\label{sec4.2}

In classification tasks like SDC, the additional outputs with XAI frameworks offer transparency by generating explanations that highlight the key characteristics and factors that influence a deep learning model to arrive at a specific class label prediction. It helps researchers diagnose the severity and spread of the disease by highlighting critical regions in medical images, providing deeper insights into the model's decision-making process.  Understanding the rationale behind a model's choice for a given input is crucial for establishing trust and ensuring accountability, particularly in medical domains like dermatology. Furthermore, the XAI framework employed in this study holds significant potential for real-world applications.

Grad-CAM helps model interpretability by offering useful insights into the relevance of features by highlighting the regions in an input image that are most influential in determining a specific classification outcome. This visualization is achieved by computing the gradient of the predicted class score concerning the feature maps in the penultimate layer. This helps researchers comprehend the areas to which priority should be given during disease treatment. The outputs of images taken for three different classes of the test dataset are shown in Figure \ref{fig10}. 

SHAP offers a complementary approach to model interpretability as it provides a comprehensive understanding of feature relevance by quantifying the importance of different patches or positions within an image. Considering various combinations of patches or tokens and assessing their impact on the prediction decisions is invaluable, especially in medical image analysis, enabling researchers and medical practitioners to pinpoint the regions or features in an image most relevant to diagnosis and treatment. The SHAP outputs of the same three images as GradCAM are shown in Figure \ref{fig11}. These Grad-CAM and SHAP insights align with quantitative results. DinoV2-B emerges as the top performer, particularly on the raw dataset, achieving a remarkable test accuracy of 96.48\%. While Swin Transformers and ViT compete closely, they fall short of DinoV2-B's performance standards.

DinoV2's Grad-CAM heatmaps and SHAP plots on the unaugmented dataset exhibit remarkable accuracy, effectively highlighting infected regions such as the hands, neck, and ears for the three images, respectively. This precision in localization elucidates why DinoV2 surpasses other architectures in performance. In contrast, its Grad-CAM heatmaps and SHAP plots on the augmented dataset show reduced accuracy due to the introduced variations, causing overfitting and impacting the efficacy of the SSL approach. The patch area is more diversified, suggesting that the model cannot narrow down to the region of infection as precisely as the model trained on the unaugmented data. These results are closely followed by the Swin Transformers model trained on the unaugmented data, with a similar area but with less intensity in and around the infected region. Nevertheless, Swin transformers, like DinoV2, demonstrate better Grad-CAM and SHAP regions on the unaugmented dataset, indicating that sliding kernel self-attention is capable of extracting relevant regions from the image.

Notably, ViT defies the norm on the augmented dataset by displaying improved Grad-CAM and SHAP performance compared to the unaugmented dataset. This underscores ViT's adaptability to diverse data distributions resulting from augmentation. It successfully captures pertinent features and regions, showcasing resilience to dataset variations.  Nevertheless, some of the regions of interest for the unaugmented data are outside the human body, suggesting room for improvement in performance. With the most attention to heatmap regions and SHAP points outside the skin in ViT, the model trained on the unaugmented data has the least ability to perform diagnosis, which aligns well with the performance metrics of the model.

\subsection{Comparison of results on smaller benchmark datasets}

To evaluate the robustness of the transformer models and showcase that the proposed pipeline can accurately automate the classification of a wide range of diseases by extracting relevant features from the input images, the models are also trained on smaller benchmark datasets such as Dermnet and HAM10000, which contain many samples per class for popular skin diseases.

\begin{figure}[h]
    \centering
    \includegraphics[width=\linewidth]{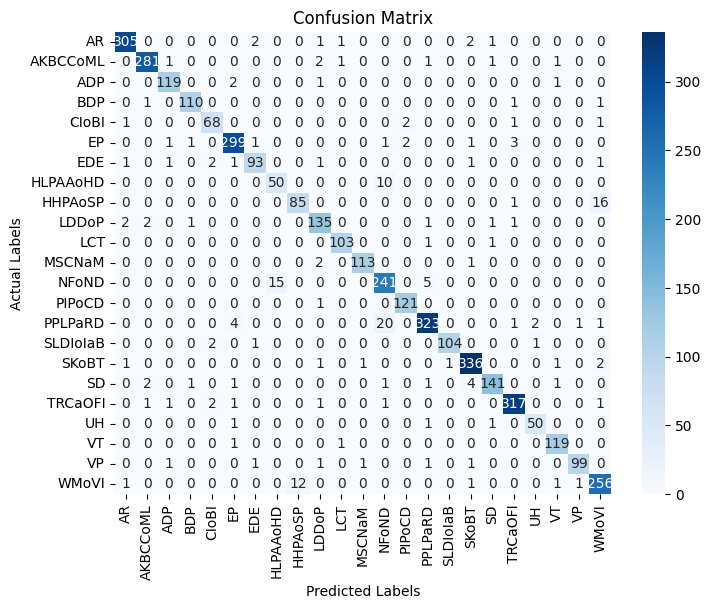}
    \caption{Confusion Matrix for the 23-class Dermnet dataset using Dino-V2.}
    \label{fig12}
\end{figure}

\begin{figure}[h]
    \centering
    \includegraphics[width=0.9\linewidth]{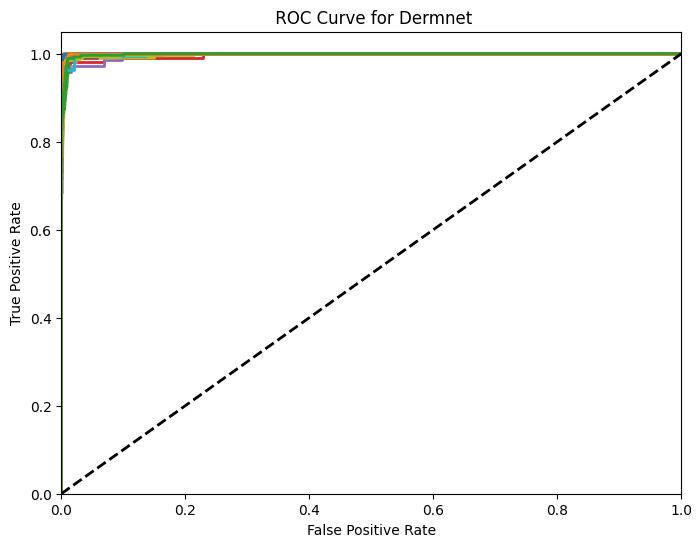}
    \caption{ROC-AUC curve for the 23-class Dermnet dataset using Dino-V2.}
    \label{fig13}
\end{figure}

\begin{table*}[h]
\centering
\caption{Comparison of SDC Models trained on the Dermnet dataset}
\label{tab4}
\begin{tabular}{|c|c|c|c|c|c|c|c|c|c||}
\hline
\textbf{Authors} & \textbf{Year} & \textbf{Classes} & \textbf{Architecture} & \textbf{Accuracy}& \textbf{Precision} & \textbf{Recall}& \textbf{F1-Score}\\
\hline
Aboulmira et al. \cite{ref42} & 2022 & 23 & DenseNet & 0.6897 & 0.6930 & 0.6920 & 0.6925 \\
Sah et. al \cite{ref43} & 2019 & 23 & Finetuned VGG & 0.7630 & 0.7600 & 0.7600 & 0.7600 \\
Bindhu et. al \cite{ref50} & 2023 & 23 & FuzzyUNet+DB& 0.9561 & \textbf{0.9472} & -  & - \\ 
Anurodh Kumar et.al \cite{ref49} & 2024 & 23 & 1D-Multiheaded CNN & 0.8857 & 0.8888 & 0.8872 & 0.8804\\ 
\hline
\textbf{Proposed Work} & & \textbf{23} & \textbf{DinoV2-Base} & \textbf{0.9623} & 0.9451 &\textbf{0.9462}	& \textbf{0.9454}\\
\hline
\end{tabular}
\end{table*}

\begin{table*}[h]
\centering
\caption{Comparison of SDC Models on the HAM10000 dataset}
\label{tab5}
\begin{tabular}{|c|c|c|c|c|c|c|c|c||}
\hline
\textbf{Authors} & \textbf{Year} & \textbf{Classes} & \textbf{Architecture} & \textbf{Accuracy}& \textbf{Precision} & \textbf{Recall}& \textbf{F1-Score}\\
\hline
Saket Chaturvedi et. al. \cite{ref28} & 2020 & 7 & ResNeXt101 & 0.9320 & 0.88 & 0.88 & 0.88\\ 
Anand et.al. \cite{ref19} & 2022 &  7 & XCeption Net & 0.9640 & - & - & - \\
Aladhadh, Suliman, et al. \cite{ref35} & 2022 & 7 & Medical-VIT & 0.9614 & \textbf{0.9600} & 0.9650 & 0.9625 \\
Krishna, Ghanta Sai et. al. \cite{ref15} & 2023 & 7 & ViT-GAN & 0.9740  & - & - & - \\
Selen Ayas \cite{ref34} & 2023 &  7 & Swin-Large & 0.9720 & 0.8510 & \textbf{0.9800} & 0.9110 \\
\hline
\textbf{Proposed Work} & & \textbf{7} & \textbf{DinoV2-Base} & \textbf{0.9745 } &0.9563	& 0.9742	&\textbf{0.9646}\\
\hline
\end{tabular}
\end{table*}

It is evident from the confusion matrix in Figure \ref{fig12} that most of the samples are being accurately classified while only a few samples from the minority classes in the dataset, such as NFoND, WMoVI, PPLRaRD and HHPAoSP (expansions in the Appendix Section \ref{sec7}). Despite the anomalies, the score for ROC-AUC curves plotted for the true positive rate vs. the false positive in Figure \ref{fig13} never falls below 0.9980. The maximum samples falling on the trace of the matrix and the lesser number of false positives and false negatives clearly demonstrate the high accuracies and F1-score indicated in Table \ref{tab4}. The table also compares the results of DinoV2-B trained on the unaugmented Dermnet dataset to other state-of-the-art works in the literature. DinoV2-B significantly outperforms the works in the literature, with an improvement in the test accuracy compared to CNN architectures.  A similar improvement can be seen in the recall and F1 scores; an overall drop in total misclassified samples improves the overall performance. 

\begin{figure}
    \centering
    \includegraphics[width=\linewidth]{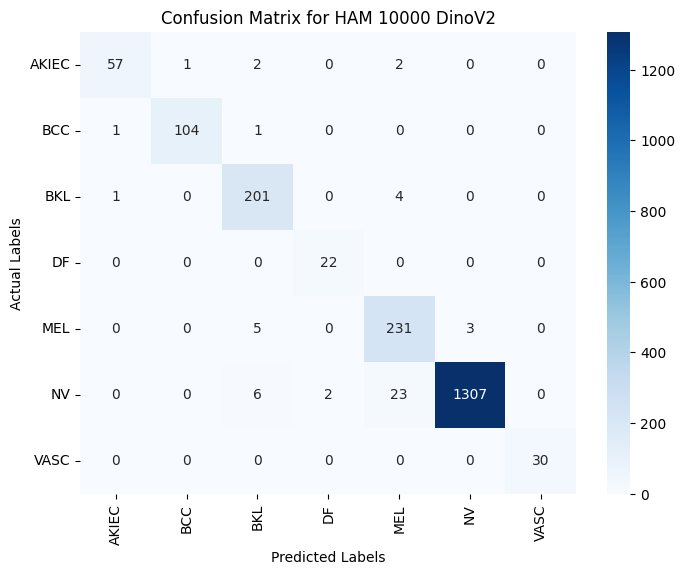}
    \caption{Confusion Matrix for the 7-class HAM10000 dataset using Dino-V2.}
    \label{fig14}
\end{figure}

\begin{figure}
    \centering
    \includegraphics[width=\linewidth]{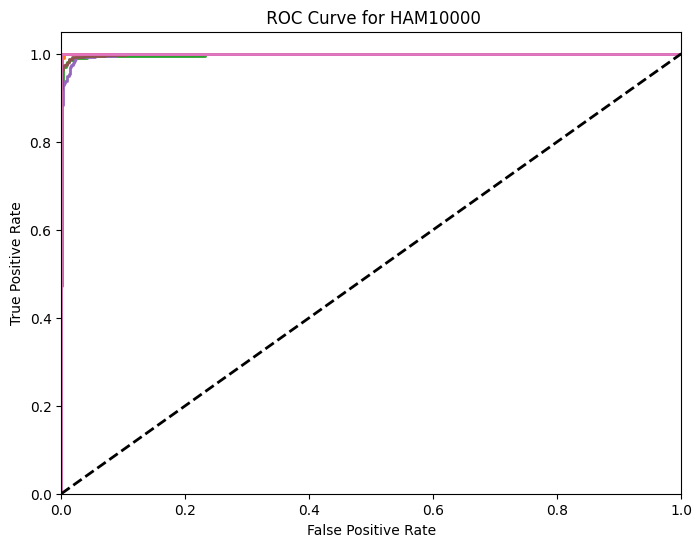}
    \caption{ROC-AUC curves for the 7-class HAM10000 dataset using Dino-V2.}
    \label{fig15}
\end{figure}

A similar trend can be seen in Table \ref{tab5}, which compares the results of DinoV2-B trained on the unaugmented HAM10000 dataset to other state-of-the-art works in the literature, with the present benchmark accuracy being a computationally heavy ViT-GAN architecture. DinoV2-B has shown a slight increase in metrics such as F1-Score, as it maintains a good balance between the precision and recall scores, which are the current state-of-the-art ones maintained by ViT-GAN and Swin-L. 

Even for highly underrepresented samples from the DF class in Figure \ref{fig14}, the model robustly classifies all test samples except 2 correctly. For the VASC class, the model accurately classifies all samples, once again proving its robustness. Yet, samples of the NV class are falsely classified as MEL due to overlapping features, which even clinicians fail to identify in extreme cases(expansions in
the Appendix Section \ref{sec7}). Nevertheless, the scores for the ROC-AUC curves in Figure \ref{fig15} are still higher than 0.995 for all classes.

These benchmark results set by DinoV2-B for both the smaller datasets in the literature with fewer samples per class are the new state-of-the-art results for small datasets probing the robustness of the transformer architectures for the SDC task.

\section{Limitations and Future work}\label{sec5}

While our proposed architecture achieved state-of-the-art results on the combined dataset, Dermnet, and HAM10000 datasets, several limitations and areas for future improvement remain to be addressed.
\begin{itemize}
    \item Computational Complexity: One of the major limitations of our methodology is its computational intensity. The models developed in this study require significant computational resources, which may not be feasible for deployment in real-time or on resource-constrained devices. Future research efforts should focus on other lightweight architectures to reduce computational complexity while maintaining or even enhancing classification performance.
    \item Generalizability to Diverse Datasets: While we demonstrated the effectiveness of our methodology on Dermnet and HAM10000 datasets, its generalizability to other diverse skin disease datasets remains unexplored. Future studies should evaluate our approach on a wider range of datasets to assess its robustness and generalizability across different skin conditions.
\end{itemize}

\section{Conclusion} \label{sec6}

The current study presented transformers to classify a diverse set of 31 skin ailments, and the results are validated with metrics like accuracy and F1-score on the data. When assessed on testing data, the final model achieved 96.48\% accuracy in detecting the condition, approximately 10\% improvement over the existing state-of-the-art results. Ten augmentation strategies were employed to improve the data distribution and determine any performance improvements, and augmentation did help CNNs like ConvNeXt and transformers like ViT to improve their performance. However, there is a drop in performance metrics SSL and sliding kernel attention techniques employed in transformers like DinoV2 and Swin transformers. According to the study results and the interpretation of them with XAI frameworks like GradCAM and SHAP, the suggested model can assist society by allowing clinicians to detect skin problems more precisely and rapidly. The improvement in results using transformers and recently-introduced DinoV2-B was also compared with other state-of-the-art results in the literature, and an improvement was observed for the 23-class Dermnet and the 7-class HAM10000 dataset. An improved recall in both datasets suggests that the improved precision suggests DinoV2-B is a robust model and can yield a lesser number of false negative diagnoses in the future. The models publically made available through this work can result in quicker and more effective therapy offered by skin specialists, enhancing patient well-being while reducing the financial burden on the healthcare system. The proposed methodology can also help the general people directly diagnose and gain an immaculate understanding of their dermatological issues without the assistance of doctors in non-complicated scenarios.

\section{Funding}
Not applicable.

\section{Conflicts of interest}
The authors declare no conflict of interest.

\section{Compliance with Ethical Standards}
None.

\section{Credit Statement}

Jayanth Mohan and Arrun Sivasubramanian – Conceptualization, Methodology, Software, Writing- Original Draft.

Sowmya V – Validation, Writing - Review \& Editing, 

Vinayakumar Ravi - Writing – Supervision.

\section{Availability of data and Code}
The data that support this study's findings are available from the first authors upon reasonable request. The best model for this dataset is deployed publically on the Hugging Face community (\href{https://huggingface.co/Jayanth2002/dinov2-base-finetuned-SkinDisease}{DinoV2-Base}, \href{https://huggingface.co/Jayanth2002/swin-base-patch4-window7-224-rawdata-finetuned-SkinDisease}{Swin-Base}, and  \href{https://huggingface.co/Jayanth2002/vit_base_patch16_224-finetuned-SkinDisease}{VIT-Base}).

\section{Appendix} \label{sec7}

\begin{table}[h]
\centering
\caption{Abbreviations and their meanings for Dermnet Dataset}
\begin{tabular}{|c|c|}
\hline
\textbf{Abbreviation} & \textbf{Meaning}                                                                                  \\ \hline
AR                    & Acne and Rosacea Photos                                                                           \\
AKBCCoML & \begin{tabular}[c]{@{}c@{}}Actinic Keratosis Basal Cell Carcinoma \\ and other Malignant Lesions\end{tabular} \\
ADP                   & Atopic Dermatitis Photos                                                                          \\
BDP                   & Bullous Disease Photos                                                                            \\
CIoBI                 & \begin{tabular}[c]{@{}c@{}}Cellulitis Impetigo and other\\  Bacterial Infections\end{tabular}     \\
EP                    & Eczema Photos                                                                                     \\
EDE                   & Exanthems and Drug Eruptions                                                                      \\
HLPAAoHD              & \begin{tabular}[c]{@{}c@{}}Hair Loss Photos Alopecia and \\ other Hair Diseases\end{tabular}      \\
HHPAoSP               & Herpes HPV and other STDs Photos                                                                  \\
LDDoP                 & Light Diseases and Disorders of Pigmentation                                                      \\
LCT                   & Lupus and other Connective Tissue diseases                                                        \\
MSCNaM                & Melanoma Skin Cancer Nevi and Moles                                                               \\
NFoND                 & Nail Fungus and other Nail Disease                                                                \\
PIPoCD                & Poison Ivy Photos and other Contact Dermatitis                                                    \\
PPLPaRD               & \begin{tabular}[c]{@{}c@{}}Psoriasis pictures Lichen Planus and \\ related diseases\end{tabular}  \\
SLDIoIaB              & \begin{tabular}[c]{@{}c@{}}Scabies Lyme Disease and\\  other Infestations and Bites\end{tabular}  \\
SKoBT                 & Seborrheic Keratoses and other Benign Tumors                                                      \\
SD                    & Systemic Disease                                                                                  \\
TRCaOFI               & \begin{tabular}[c]{@{}c@{}}Tinea Ringworm Candidiasis and \\ other Fungal Infections\end{tabular} \\
UH                    & Urticaria Hives                                                                                   \\
VT                    & Vascular Tumors                                                                                   \\
VP                    & Vasculitis Photos                                                                                 \\
WMoVI                 & Warts Molluscum and other Viral Infections                                                        \\ \hline
\end{tabular}
\label{tab6}
\end{table}

\begin{table}[h]
    \centering
     \caption{Abbreviations and their meanings for HAM10000\\  Dataset}

    \begin{tabular}{|c|c|}
\hline
\textbf{Abbreviation} & \textbf{Meaning}                                \\ \hline
AKIEC                 & Actinic Keratoses and Intraepithelial Carcinoma \\
BCC                   & Basal Cell Carcinoma                            \\
BKL                   & Benign Keratosis Like Lesions                   \\
DF                    & Dermatofibroma                                  \\
MEL                   & Melanoma                                        \\
NV                    & Melanocytic Nevi                                \\
VASC                  & Vascular Lesions                                \\ \hline
\end{tabular}
    \label{tab:abbreviations ham}
\end{table}

\end{document}